\documentclass[letterpaper, 10 pt, conference]{ieeeconf}
\IEEEoverridecommandlockouts
\usepackage{cite}
\usepackage{amsmath,amssymb,amsfonts}
\usepackage{graphicx}
\usepackage{textcomp}
\usepackage{xcolor}
\usepackage{svg}
\usepackage{lipsum}
\usepackage[thinc]{esdiff}
\usepackage{mathtools}
\usepackage{cleveref}
\usepackage{textcomp}
\usepackage{tabularx}
\usepackage{siunitx}
\usepackage{caption}
\usepackage{subcaption}
\usepackage{upgreek}
\usepackage{stackengine}
\renewcommand*\descriptionlabel[1]{\hspace\leftmargin$#1$}

\title{\LARGE \bf Ballistic Multibody Estimator for 2D Open Kinematic Chain}

\author{Thanacha Choopojcharoen$^{1,2,\ddag}$, Worachit Ketrungsri$^{1,\ddag}$, \\Thanapong Chuangyanyong$^{1}$, and Panusorn Chinsakuljaroen$^{1}$
\thanks{$^{\ddag}$These authors contributed equally to this work.}%
\thanks{$^{1}$All authors are associated with the Institute of Field Robotics, King Mongkut's University of Technology Thonburi, 126 Pracha Uthit Rd., Bang Mot, Thung Khru, Bangkok 10140.}%
\thanks{$^{2}$Thanacha Choopojcharoen is currently an adjunct lecturer at the Institute of Field Robotics, King Mongkut's University of Technology Thonburi.}%
}

\begin{document}

\maketitle

\thispagestyle{empty}

\pagestyle{empty}

\begin{abstract}
Applications of free-flying robots range from entertainment purposes to aerospace applications. The control algorithm for such systems requires accurate estimation of their states based on sensor feedback. The objective of this paper is to design and verify a lightweight state estimation algorithm for a free-flying open kinematic chain that estimates the state of its center-of-mass and its posture. Instead of utilizing a nonlinear dynamics model, this research proposes a cascade structure of two Kalman filters (KF), which relies on the information from the ballistic motion of free-falling multibody systems together with feedback from an inertial measurement unit (IMU) and encoders. Multiple algorithms are verified in the simulation that mimics real-world circumstances with Simulink. Several uncertain physical parameters are varied, and the result shows that the proposed estimator outperforms EKF and UKF in terms of tracking performance and computational time.
\end{abstract}

\section{Introduction}
The development of free-flying systems is a growing area in robotics. There are numerous applications of this field, such as space manipulators, hopping robots, and acrobatic robots. \cite{dung-2017} built a three-link gymnastics robot that can perform a front-flip motion. \cite{pope-2018} presented a two-degree-of-freedom robot, which produces somersaulting stunts. Papadopoulos and Fragkos \cite{papadopoulos-2007} developed a planning algorithm for a $n$-link gymnastics robot with non-zero initial angular momentum. In development, most of them usually utilize external positioning sensors, such as cameras or motion capture systems. This limits the system to stay in the indoor setting or laboratory. However, when deployed, if such systems need to be self-contained, systems must obtain their states without the help of external positioning sensors. One of the most prevalent and cost-effective implementations is the use of an inertial measurement unit (IMU) together with a nonlinear state estimator. A prominent example of such a state estimator is the extended Kalman filter (EKF) \cite{prevost-2007}, which uses linearized system dynamics as the process model. However, EKF tends to perform inadequately on a highly nonlinear system \cite{barrau-2020}. Hence, several approaches were made to overcome these problems. One of the aforementioned is to use the unscented Kalman filter (UKF), which applies a deterministic sampling approach that makes the estimation more adaptive. Thus, the resulting estimation error is usually less than the traditional EKF with higher computational time \cite{kotecha-2003}. In general, an inaccurate dynamic model of the system of interest often leads to unreliable estimation. 

This research focuses on designing and verifying a lightweight estimation algorithm for a free-flying open kinematic chain. This paper proposed a new cascaded state estimation algorithm called ``Ballistic Multibody Estimator" (BME), which is capable of estimating the state of the center-of-mass (position and velocity) and the state of joints (position and velocity) of an open kinematic chain by using only the Kalman filter (KF) \cite{kalman-1960} along with input measurements from an IMU and encoders. Consequently, this algorithm can be implemented in low-performance embedded systems; thereby, offering a cost-effective solution.
 
This algorithm is applied to an accurate model of  Mon$\chi$, a two-link free-flying acrobatic robot. The measurement signals used in this paper are simulated from the system's encoders and IMU. No additional external positioning sensors are needed since this configuration is sufficient for the algorithm. Both EKF and UKF are also studied and compared with the proposed estimator in the simulation that mimics real-world circumstances.

\section{Problem Formulation}
This research focuses on designing a state estimation algorithm for $n$-link open kinematic chain in 2D space. Each link is connected in series by revolute joints with a parallel axis of rotation. The equation of motion is derived in the same manner as a serial manipulator. Although this paper does not use a generic open kinematic chain, the kinematics and dynamics model of the generic system can be computed \cite{featherstone-2007} and systematically applied to the proposed algorithm.

\subsection{System Description}
\begin{figure}[htbp]
    \centerline{\includegraphics[width=\textwidth/2]{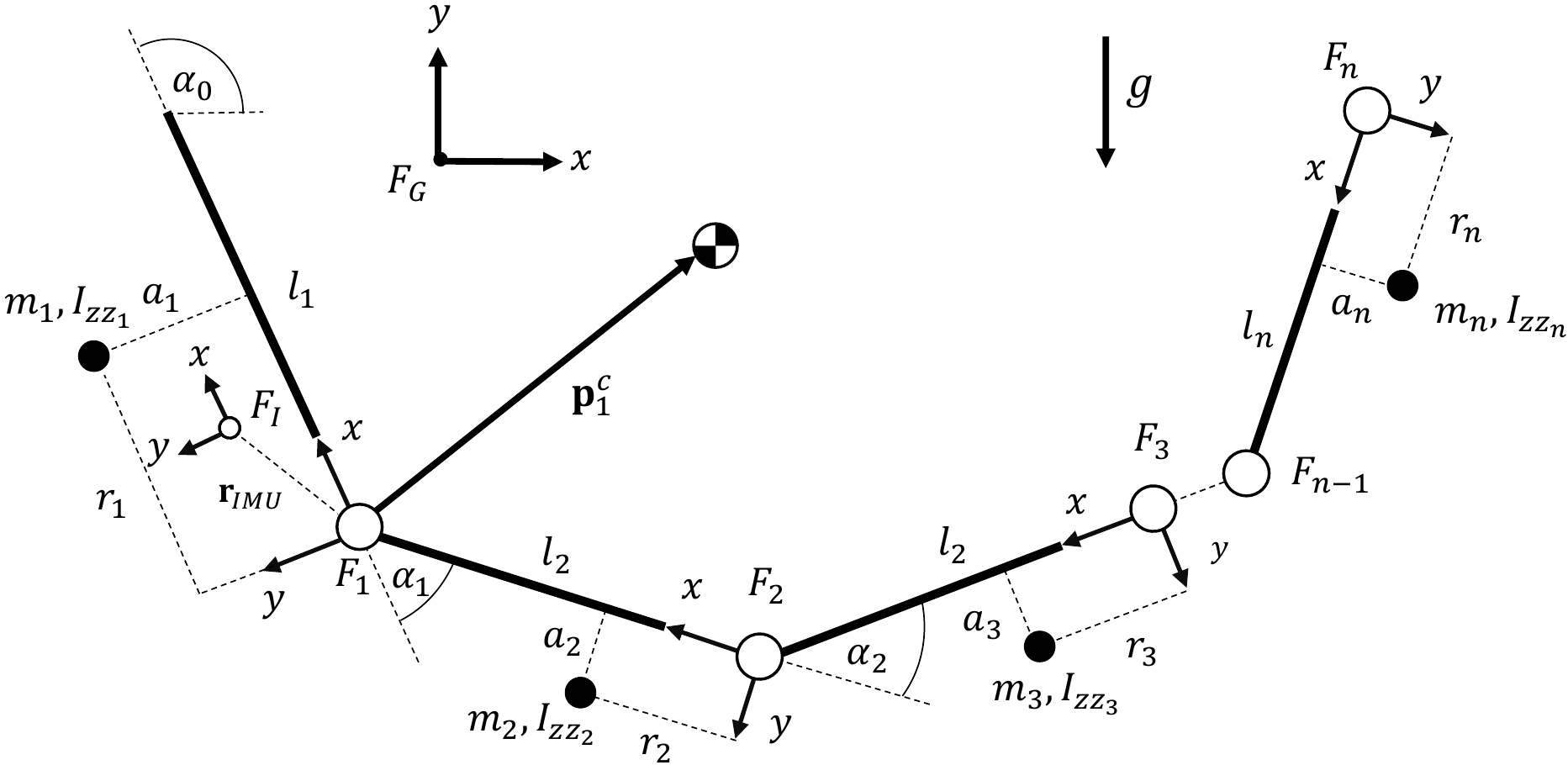}}
    \caption{Open kinematic chain model.}
    \label{flying_multibody}
\end{figure}

Fig.~\ref{flying_multibody} shows the model of the robot. The flying open kinematic chain has $n$ links and $n-1$ joints. A global coordinate frame is denoted by $F_G$. A body-attached coordinate frame of link $i$ is denoted by $F_i$, where $i \in \{1,2,\dots,n\}$. Joint $i$ is the joint that is connected between link $i$ and link $i+1$, which has an actuation effort of $\tau_i$. Each joint is modeled to exhibit linear damping with a damping coefficient of $\beta_i$.
The x-axis of the body-attached coordinate frame aligns with the origin of $F_{i-1}$ and $F_i$ and points toward $F_{i-1}$. Its origin coincides with the axis of rotation of the joint $i$ and its x-axis. The mass and the moment of inertia of each link with respect to the local frame ($F_i$) are denoted by $m_i$ and $I_{{zz}_i}$, respectively. The distance of the center-of-mass (CoM) of each link along the x-axis and y-axis of $F_i$ are denoted by $a_i$ and $r_i$, respectively. The IMU's coordinate frame is denoted by $F_I$. $\mathbf{r}_\text{IMU}$ is the position of the origin of the $F_I$ relative to $F_1$. The posture of a 2D open kinematic chain ($\bar{\mathbf{q}}$) refers to the orientation of the first link ($\alpha_0$), which is the angle between the x-axis of $F_1$ and the x-axis of $F_G$; and the angle of joint $i$ ($\alpha_i$) with respect to the x-axis of $F_i$ \eqref{eq:posture}.  

\begin{equation}\label{eq:posture}
\bar{\mathbf{q}}=
\begin{bmatrix}
    \alpha_0 \\ \alpha_1 \\ \vdots \\ \alpha_{n-1}
\end{bmatrix}
\end{equation} 

Since this paper analyzes a system in the 2D space, to ensure that the multibody system remains on the xy-plane, the moment of inertia of the whole body taken at the center-of-mass align with the body attached frame ($\mathbf{I}_c$) must take the form in \eqref{eq:I_sagittal}.

\begin{equation}\label{eq:I_sagittal}
    \mathbf{I}_c = 
    \begin{bmatrix}
        I_{xx} & I_{xy} & 0 \\
        I_{yx} & I_{yy} & 0 \\
        0      & 0      & I_{zz} \\
    \end{bmatrix}
\end{equation}

\subsection{Sensor Setup}
A 6-DOF IMU and absolute encoders are mounted on the robot. The measurement from both sensors is modeled as an additive zero-mean Gaussian white noise. The IMU is mounted on the first link of the robot such that $F_I$ is aligned with $F_1$. Furthermore, the IMU comprises a gyroscope and an accelerometer; the former measures the angular velocity of the object. The latter measures the specific force response of the attached object. Even though the sensor has 6-DOF, the measurement only used 3-DOF since the system stays in the 2D plane. These quantities consist of angular velocity in the z-axis, the specific force that acts upon the x-axis and y-axis of $F_I$. The gyroscope's measurement ($z_g$) can be modeled in terms of the angular velocity of the first link \eqref{eq:math_gyro}.

\begin{equation}\label{eq:math_gyro}
    z_g =  \dot{\alpha}_0
\end{equation}

 The measurement from the accelerometer ($\mathbf{z}_a$), in \eqref{eq:math_accel}, is based on the specific force, which can be modeled in terms of acceleration of the origin of $F_I$ with respect to $F_G$ ($\ddot{\mathbf{p}}^I_G$).  

\begin{equation}\label{eq:math_accel}
    \mathbf{z}_a =  \mathbf{R}(\alpha_0)^\top\left(\ddot{\mathbf{p}}^I_G - \begin{bmatrix}
        0 & -g
    \end{bmatrix}\right)^\top
\end{equation}

$g$ is the gravitational acceleration constant. $\mathbf{R} \in \mathbf{SO}(2)$ is the rotation matrix that represents the orientation of the $F_I$ relative to $F_G$, which depends only on $\alpha_0$.

The last sets of sensors are the absolute encoder, which is mounted on each joint. Each of them provides measurements ($\mathbf{z}_{e,i}$), in \eqref{eq:math_encoders}, that correspond to angular position ($\alpha_i$) and angular velocity ($\dot{\alpha}_i$) of the corresponding joint.

\begin{equation}\label{eq:math_encoders}
    \mathbf{z}_{e,i} =  \begin{bmatrix}
        \alpha_i \\ \dot{\alpha}_i
    \end{bmatrix}
\end{equation}

\section{Kalman Filters with Full Dynamics}
Based on the system description, the generalized coordinate ($\mathbf{q}$) consists of $n+2$ components \eqref{eq:generalized_coordinate}, which categorized into the posture and the overall position. 
One way to describe the position is to choose a reference point on one of the links.
In this paper, we choose $d$ and $h$, which are the distance from the origin of $F_G$ to the origin of $F_1$.

\begin{equation}\label{eq:generalized_coordinate}
\mathbf{q} =
    \begin{bmatrix}
        \bar{\mathbf{q}} \\ d \\ h
    \end{bmatrix}
\end{equation}

\subsection{Full Dynamics Model}
 From \eqref{eq:generalized_coordinate}, the dynamics model of 2D rigid body dynamics can be written as seen in \eqref{eq:ss_eq}. 
 The generalized mass and inertia matrix ($\mathbf{M}$) , the Coriolis and centrifugal  terms ($\mathbf{C}$), and the  gravitational  terms ($\mathbf{g}$) can  be  derived  from  the linear Jacobian ($\mathbf{J}_{v_i}$) and angular Jacobian ($\mathbf{J}_{\omega_i}$) according to \cite{featherstone-2007}. 
 The matrix of damping coefficient ($\mathbf{B}$) is defined by \eqref{eq:B} since there is only $n-1$ joints.

\begin{equation}\label{eq:ss_eq}
    \ddot{\mathbf{q}} =  \mathbf{a}(\mathbf{q},\dot{\mathbf{q}},\mathbf{u}) = 
    \mathbf{M}^{-1}\left(
        \mathbf{u}
        -\mathbf{C}\dot{\mathbf{q}}-\mathbf{B}\dot{\mathbf{q}}-\mathbf{g}\right)
\end{equation}

Where:

\begin{equation*}
    \mathbf{M} = \sum_{i=1}^n m_i\mathbf{J}^\top_{v_i}\mathbf{J}_{v_i} +  I_{{zz}_i}\mathbf{J}^\top_{\omega_i}\mathbf{J}_{\omega_i}
\end{equation*}

\begin{equation*}
    C_{ij}=\frac{1}{2}\sum_{k=1}^{n+2}\bigg(\frac{\partial M_{ij}}{\partial q_k }+\frac{\partial M_{ik}}{\partial q_j}-\frac{\partial M_{kj}}{\partial q_i}\bigg)q_k
\end{equation*}

\begin{equation*}
    \mathbf{g} = \sum_{i=1}^n\big(- m_i\begin{bmatrix}
        0 & -g
    \end{bmatrix}\mathbf{J}_{v_i}\big)^\top
\end{equation*}

\begin{equation}\label{eq:B}
    \mathbf{B} = \text{diag}(0,\beta_1,\beta_2,\cdots,\beta_{n-1},0,0)
\end{equation}

$\text{diag} (\cdot)$ is a diagonal matrix with the arguments in the diagonal components, respectively. The linear Jacobian \eqref{eq:J_v} is derived from a partial derivative of a position vector from the global-coordinate frame to the CoM of each link ($\mathbf{p}^{c_i}_G$) with respect to the configuration variable. The linear Jacobian matrix \eqref{eq:J_omega} is specified by the entry in the $j$-th column, where $\mathbf{J}_{\omega_i} \in \mathbb{R}^{1 \times (n+2)}$.

\begin{equation}\label{eq:J_v}
    \mathbf{J}_{v_i} 
    = 
    \frac{\partial\mathbf{p}^{c_i}_G}{\partial \mathbf{q}}
    =  
    \frac{\partial} {\partial \mathbf{q}} \left(
    \begin{bmatrix}
        d \\ 
        h
    \end{bmatrix} 
    +
    \mathbf{p}^{c_i}_1(\bar{\mathbf{q}})\right)
\end{equation}

\begin{equation}\label{eq:J_omega}
    \mathbf{J}_{\omega_{i,j}} = \begin{cases*}
      1, & if $j \leq i$,\\
      0,                    & otherwise.
    \end{cases*}
\end{equation}
Where:
\begin{equation}\label{eq:p_c1}
    \mathbf{p}^{c_i}_1(\bar{\mathbf{q}}) =
    \mathbf{p}_1^i+\mathbf{R}\left(\sigma_i\right)
    \begin{bmatrix}
        r_i\\a_i
    \end{bmatrix}
    ,~~
    \sigma_\lambda = \sum_{k=0}^{\lambda-1} \alpha_k
\end{equation}

\begin{equation}
    \mathbf{p}_1^i = 
    \begin{cases}
        \mathbf{0}_{2\times 1}, & i=1 \\
        \sum_{j=2}^{i}l_{j}
        \begin{bmatrix}
          \cos(\sigma_j)\\
          \sin(\sigma_j)
        \end{bmatrix}, & \text{otherwise.} 
    \end{cases}
\end{equation}

Note that $\mathbf{p}^{c_i}_0$ refers to the position of CoM relative to the origin of $F_0$.
The generalized control input, $\mathbf{u}$, comprises the joint efforts, $\tau_i$, for the first $n$ components and zeros for the last two components, where $\mathbf{u}\in\mathbb{R}^{(n+2)}$.

\subsection{Equations for Estimators}
In this paper, the structure of discrete-time EKF and UKF refers to \cite{thrun-2005} and \cite{wan-2000} with the estimate states ($\mathbf{x}$) that comprises $\mathbf{q}$ and $\dot{\mathbf{q}}$.

\begin{equation}
    \mathbf{x} =
    \begin{bmatrix}
        \mathbf{q} \\ \dot{\mathbf{q}} 
    \end{bmatrix}
\end{equation}

The predicted state ($\mathbf{x}^*$) can be computed from the discretized nonlinear prediction model in the form of

\begin{equation}\label{eq:discretized}
    \mathbf{x}^*[k] = \mathbf{x}[k-1] + 
             \Delta t \cdot
             \begin{bmatrix}
            \dot{\mathbf{q}}[k-1] \\
            \mathbf{a}(\mathbf{q}[k-1],\dot{\mathbf{q}}[k-1],\mathbf{u}[k])
            \end{bmatrix}
\end{equation}

where $\Delta t$ denotes the time-step of the estimator. 

According to \cite{wan-2000,thrun-2005}, the state transition matrix $\mathbf{F}[k]$ must also be provided for predicting the state covariance matrix, which can be computed by taking the partial derivative of \eqref{eq:discretized} with respect to $\mathbf{x}[k-1]$ and evaluate at $\mathbf{x}[k-1]$ and $\mathbf{u}[k]$. Once the state is predicted using \eqref{eq:discretized}, the result is used to obtain the predicted the measurement ($\mathbf{y}$) based on the sensor models \eqref{eq:math_gyro}, \eqref{eq:math_accel}, and \eqref{eq:math_encoders}. 

\begin{equation}\label{eq:sensors_model}
    \mathbf{y}[k] = \mathbf{h}(\mathbf{x}^*[k],\mathbf{u}[k])
\end{equation}
Where:
\begin{equation*}
\mathbf{h}(\mathbf{x},\mathbf{u}) = 
\begin{bmatrix}
        z_g \\ \mathbf{z}_a \\ \mathbf{z}_{e,1} \\ \vdots \\  \mathbf{z}_{e,n}
    \end{bmatrix}
\end{equation*}

The output matrix ($\mathbf{H}$) is approximated by taking the partial derivative of \eqref{eq:sensors_model} and evaluating the derivative at the predicted state $\mathbf{x}^*[k]$ and the control input $\mathbf{u}[k]$. Note that the partial derivatives in $\mathbf{F}[k]$ and $\mathbf{H}[k]$ are difficult to obtain in a general form. Thus, an adjustment in the model needs to be made. 

\section{Kalman Filters with Reduced Dynamics}
Another way to describe the overall position of a free-flying open kinematic system is using its CoM. Moreover, since there is no external force acting on the CoM, a posture of the flying kinematic chain is independent of the position and velocity of its CoM. Hence, the dynamics model can be decoupled between its CoM and its posture. This results in a new architecture of state estimation. 

\subsection{Reduced Dynamics Model}
According to \cite{kleppner-2010}, the CoM will follow a parabolic trajectory, which implies that a kinematic model of a 2D ballistic motion of a free-fall body can be used to describe the behavior of the CoM with respect to $F_G$, which evolves independently from the posture of the robot. Thus, a reduced dynamics model can be obtained independently of the CoM states, which only consists of the posture. The corresponding reduced generalized coordinate is only the posture ($\bar{\mathbf{q}}$). According to \cite{dubowsky-1993}, a free-flying kinematic chain can be modeled with respect to a frame that shares an origin with its CoM and is aligned with $F_G$ since the coordinate frame is inertial. This frame is denoted by $F_c$. The dynamics model of the reduced equation of motion \eqref{eq:manipulator_eq_bar} can be derived similarly to \eqref{eq:ss_eq}.
    
\begin{equation}\label{eq:manipulator_eq_bar}
    \ddot{\bar{\mathbf{q}}} =  \bar{\mathbf{a}}(\bar{\mathbf{q}},\dot{\bar{\mathbf{q}}},\bar{\mathbf{u}}) = \bar{\mathbf{M}}^{-1}\left(
        \bar{\mathbf{u}}
        -\bar{\mathbf{C}}\dot{\bar{\mathbf{q}}}-\bar{\mathbf{B}}\dot{\bar{\mathbf{q}}}-\bar{\mathbf{g}}
        \right)
\end{equation}

The reduced mass and inertia matrix ($\bar{\mathbf{M}}$), reduced Coriolis matrix ($\bar{\mathbf{C}}$), and reduced gravitational terms ($\bar{\mathbf{g}}$) can be computed from the reduced linear Jacobian matrix $(\bar{\mathbf{J}}_{v_i})$ in \eqref{eq:J_v2} and the reduced angular Jacobian matrix ($\bar{\mathbf{J}}_\omega$) in \eqref{eq:J_omega2}.
    
\begin{equation}\label{eq:J_v2}
    \bar{\mathbf{J}}_{v_i} =         \frac{\partial\mathbf{p}^{c_i}_c} {\partial     \bar{\mathbf{q}}}=
    \frac{\partial}{\partial\bar{\mathbf{q}}}\left(\mathbf{p}^{c_i}_1(\bar{\mathbf{q}}) - \mathbf{p}_1^c \right)
\end{equation}
    
\begin{equation}\label{eq:J_omega2}
    \bar{\mathbf{J}}_{\omega_{i,j}} = \begin{cases*}
      1, & if $j \leq i$,\\
      0,                    & otherwise.
    \end{cases*}
\end{equation}
Where:
\begin{equation}\label{eq:local_com}
    \mathbf{p}_1^c = \frac{\sum_{j=1}^n m_j \mathbf{p}_1^{c_j}(\bar{\mathbf{q}})}{\sum_{j=1}^n m_j}
\end{equation}

Note that $\mathbf{p}^{c_i}_c$ is the position of the CoM of the link $i$ relative to the origin of $F_c$, which can be computed by \eqref{eq:p_c1}.
The reduced matrix of damping coefficient is similar to \eqref{eq:B}.

\begin{equation*}
    \bar{\mathbf{B}} = \text{diag}(0,\beta_1,\beta_2,\cdots,\beta_{n-1})
\end{equation*}

Because the generalized coordinate is reduced, the reduced generalized control input ($\bar{\mathbf{u}}$) consists of the joint efforts $\tau_i$ in all components, where $\bar{\mathbf{u}}\in\mathbb{R}^{n}$.

\subsection{A new cascade architecture for estimation}
The idea is to estimate the posture separately from the CoM. For this architecture to work, the posture estimation must estimate the acceleration of all joint angles ($\ddot{\bar{\mathbf{q}}}$), which later be used for CoM estimation. Therefore, the state of posture for posture estimation ($\mathbf{x}_p$) is defined to be \eqref{eq:x_p}.

\begin{equation}\label{eq:x_p}
    \mathbf{x}_p = 
    \begin{bmatrix}
            \bar{\mathbf{q}} \\ \dot{\bar{\mathbf{q}}} \\ \ddot{\bar{\mathbf{q}}}
    \end{bmatrix}
\end{equation}

This affects the structure of the discretized nonlinear prediction model, which can be expressed in the form of \eqref{eq:reduced_discretized}, where $\bar{\mathbf{x}}_p$ is the predicted reduced state. The state transition matrix for posture estimation $\mathbf{F}_p[k]$ is still computed in the same fashion  by taking the partial derivative of \eqref{eq:reduced_discretized} with respect to $\mathbf{x}_p$ and evaluating the derivative at the previous estimated state $\mathbf{x}_p[k-1]$ and the control input $\bar{\mathbf{u}}[k]$.

\begin{equation}\label{eq:reduced_discretized}
    \mathbf{x}_p^*[k] =  
    \begin{bmatrix}
            \bar{\mathbf{q}}[k-1]+\Delta t\cdot \dot{\bar{\mathbf{q}}}[k-1]\\
            \dot{\bar{\mathbf{q}}}[k-1]+\Delta t\cdot \bar{\mathbf{a}}(\bar{\mathbf{q}}[k-1],\dot{\bar{\mathbf{q}}}[k-1],\bar{\mathbf{u}}[k]) \\
            \bar{\mathbf{a}}(\mathbf{\bar{q}}[k-1],\dot{\bar{\mathbf{q}}}[k-1],\bar{\mathbf{u}}[k])
    \end{bmatrix}
\end{equation}

In this architecture, only the gyroscope and encoders are used for estimating the posture. The predicted the measurement for posture estimation ($\mathbf{y}_p$) are based on the sensor models \eqref{eq:math_gyro} and \eqref{eq:math_encoders}. 

\begin{equation}\label{eq:posture_sensors_model}
    \mathbf{y}_p[k] = \mathbf{h}_p(\mathbf{x}_p^*[k],\bar{\mathbf{u}}[k])  
\end{equation}

Where:
\begin{equation*}
\mathbf{h}_p(\mathbf{x}_p,\bar{\mathbf{u}}) = 
\begin{bmatrix}
        z_g \\ \mathbf{z}_{e,1} \\ \vdots \\  \mathbf{z}_{e,n}
    \end{bmatrix}
\end{equation*}

The observation matrix for posture estimation ($\mathbf{H}_p[k]$) is approximated by taking the partial derivative of \eqref{eq:posture_sensors_model} and evaluating at the predicted state $\mathbf{x}_p^*[k]$ and the control input $\bar{\mathbf{u}}[k]$. Since the both state and measurement prediction models are nonlinear, the posture estimator can be chosen as EKF or UKF that uses \eqref{eq:reduced_discretized} and \eqref{eq:posture_sensors_model} along with $\mathbf{F}_p[k]$ and $\mathbf{H}_p[k]$.

Separately, the CoM estimator is based on a zero-mean Gaussian jerk model. The continuous-time model in \eqref{eq:cont} can be used to compute the motion of the CoM.

\begin{equation}\label{eq:cont}
    \begin{bmatrix}
        p_x(t) \\ \dot{p}_x(t) \\ \ddot{p}_x(t) \\ p_y(t) \\ \dot{p}_y(t) \\ \ddot{p}_y(t)
    \end{bmatrix}
    =
    \begin{bmatrix}
        p_x(t_0)+\dot{p}_x(t_0)t+\frac{1}{6}j_x t^3 \\ \dot{p}_x(t_0)+\frac{1}{2}j_x t^2 \\ 
        j_x t \\
        p_y(t_0)+\dot{p}_y(t_0)t-\frac{1}{2}g t^2+\frac{1}{6}j_y t^3 \\ 
        \dot{p}_y(t_0)-gt+\frac{1}{2}j_y t^2\\
        -g+j_y t
    \end{bmatrix}
\end{equation}
Where:
\begin{equation*}
    \mathbf{p}_G^c = 
    \begin{bmatrix}
            p_x \\ p_y
    \end{bmatrix}
\end{equation*}

$j_x$ and $j_y$ denote the jerk along both the x-axis and the y-axis of $F_G$, which is assumed to be zero-mean Gaussian. The need for predicting the acceleration is due to the relationship between the kinematic and the specific force measured by the accelerometer. 
This results in the discretized kinematic-based prediction model as seen in \eqref{eq:x_k}, which is linear in terms of the state of CoM ($\mathbf{x}_c$). And $\otimes$ is the Kronecker product operator.

    \begin{equation}\label{eq:x_k}
    \mathbf{x}_c[k] = \mathbf{F}_c \mathbf{x}_{c}[k-1] + \mathbf{G}_c\mathbf{w}_c[k] +  \mathbf{u}_c 
    \end{equation}

Where: 
\begin{equation*}\label{eq:x_c}
    \mathbf{x}_c = \begin{bmatrix}
        p_x &
        \dot{p}_x &
        \ddot{p}_x &
        p_y &
        \dot{p}_y &
        \ddot{p}_y
    \end{bmatrix}^\top
\end{equation*}

\begin{equation*}\label{eq:w_c_u_c}
    \mathbf{w}_c = 
    \begin{bmatrix}
        j_x \\ j_y
    \end{bmatrix}
    ,~~
    \mathbf{u}_c = 
    \begin{bmatrix}
        \mathbf{0}_{1\times3} & -\frac{1}{2}g\Delta{t}^2 & -g\Delta{t} & -g
    \end{bmatrix}^\top
\end{equation*}
    
\begin{equation}\label{eq:F_c_G_c}
    \mathbf{F}_c = \mathbf{\mathbb{I}}_2 \otimes      \begin{bmatrix}
    1 & \Delta{t} & 0\\
    0 & 1 & 0\\
    0 & 0 & 0\\
    \end{bmatrix} 
    ,~~
    \mathbf{G}_c =  \mathbf{\mathbb{I}}_2 \otimes \begin{bmatrix}
        \frac{1}{6}\Delta{t}^3 \\
        \frac{1}{2}\Delta{t}^2 \\
        \Delta{t}
    \end{bmatrix}
\end{equation}

Being the only sensor for CoM estimation, the accelerometer measures the specific force at $F_I$. From \eqref{eq:math_accel}, the acceleration of the origin of $F_I$ relative to the origin of $F_G$ ($\ddot{p}_G^I$) can be computed by using \eqref{eq:local_com} and \eqref{eq:acc_fk}.

\begin{equation}\label{eq:acc_fk}
    \ddot{\mathbf{p}}_G^I - \ddot{\mathbf{p}}_1^I = \ddot{\mathbf{p}}_G^c - \ddot{\mathbf{p}}_1^c
\end{equation}
Where:
\begin{equation*}
\ddot{\mathbf{p}}_1^I = \diff[2]{}{t}\left(\mathbf{R}(\alpha_0)\mathbf{r}_\text{IMU}\right)
    =
    \mathbf{R}(\alpha_0)
    \mathbf{\Lambda}
    \mathbf{r}_\text{IMU}
\end{equation*}

\begin{equation*}
    \mathbf{\Lambda} = 
    \begin{bmatrix}
            - \dot{\alpha}_0^2 & -\Ddot{\alpha}_0 \\
            \Ddot{\alpha}_0 & - \dot{\alpha}_0^2 \\
    \end{bmatrix}
\end{equation*}

Note that both $\ddot{\mathbf{p}}_0^I$ and $\ddot{\mathbf{p}}_0^c$ can be computed from $\bar{\mathbf{q}}$, $\dot{\bar{\mathbf{q}}}$, and $\ddot{\bar{\mathbf{q}}}$. 
This implies that the estimated posture state ($\mathbf{x}_p$) from the posture estimator can be used to feed through the output equation of the CoM estimator, which can be expressed as \eqref{eq:output_com}. 

\begin{equation}\label{eq:output_com}
    \mathbf{y}_c[k] = \mathbf{h}_c(\mathbf{x}_c[k],\hat{\mathbf{x}}_p[k]) 
\end{equation}

Where:

\begin{equation*}\label{eq:h_c}
    \mathbf{h}_c(\mathbf{x}_c,\mathbf{x}_p) =  
    \mathbf{R}(\alpha_0)^\top\left( \begin{bmatrix}
    \ddot{p}_x \\ \ddot{p}_y+g
    \end{bmatrix} - 
    \ddot{\mathbf{p}}_1^c \right) +
    \mathbf{\Lambda}\mathbf{r}_\text{IMU}
\end{equation*}

\eqref{eq:output_com} only depends on $\hat{\mathbf{x}}_p$ and $\mathbf{p}_G^c$. Therefore, when taking the partial derivative of \eqref{eq:output_com} with respect to $\mathbf{x}_c$, the output matrix of the CoM estimator ($\mathbf{H}_c$) only depends on $\alpha_0$ as seen in \eqref{eq:H_c}. 

\begin{equation}\label{eq:H_c}
\mathbf{H}_c[k] = 
    \begin{bmatrix}
        \mathbf{0}_{2\times2} & 
        \pmb\upeta_x &
        \mathbf{0}_{2\times2} &
        \pmb\upeta_y
    \end{bmatrix}
\end{equation}

Where:

\begin{equation*}
    \pmb\upeta_\lambda = \diffp[]{\mathbf{h}_c}{\ddot{p}_\lambda}\Bigr|_{ \substack{{\bar{\mathbf{\alpha}_0}}={\alpha}^*[k]\hfill}}
\end{equation*}

$\lambda$ refers to a component of $\ddot{p}$, which can be either $x$ or $y$.

If the estimated states of posture ($\mathbf{x}_p$) are treated as a time-varying signal, \eqref{eq:output_com} can also be treated as a linear time-varying system. Hence, a time-varying KF is chosen to estimate the CoM based on \eqref{eq:x_k}, \eqref{eq:F_c_G_c}, \eqref{eq:output_com}, and \eqref{eq:H_c}. 
From \eqref{eq:acc_fk}, the estimated states of posture ($\hat{\mathbf{x}}_c$) and the estimated states of CoM ($\hat{\mathbf{x}}_p$) can also be used for computing $d$, $h$,  $\dot{d}$, and $\dot{h}$ as seen in \eqref{eq:transfrom_dh_dh_dot}.

\begin{equation}\label{eq:transfrom_dh_dh_dot}
    \begin{bmatrix}
        d \\ h
    \end{bmatrix} = 
    \begin{bmatrix}
        p_x \\ p_y
    \end{bmatrix} - 
    \mathbf{p}_1^c
    ,~~
    \begin{bmatrix}
        \dot{d} \\ \dot{h}
    \end{bmatrix} = 
    \begin{bmatrix}
        \dot{p}_x \\ \dot{p}_y
    \end{bmatrix} - 
    \dot{\mathbf{p}}_1^c
\end{equation}

This architecture will be referred to as a ``Decoupled Estimator" (DE). The CoM estimator of DE, which is a linear time-varying KF, relies on the estimated posture from the posture estimator, which can be chosen as EKF or UKF.

\section{Ballistic Multibody Estimator}
\begin{figure}[htbp]
        \centerline{\includegraphics[scale=0.6]{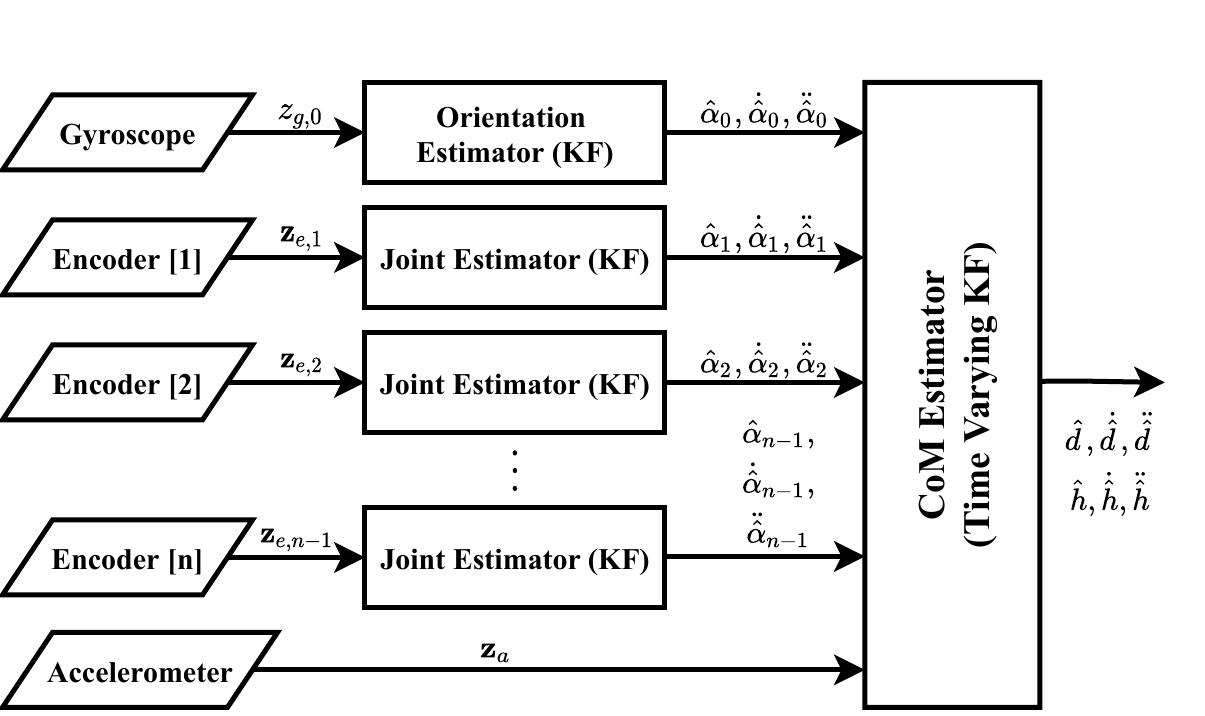}}
        \caption{Ballistic Multibody Estimator scheme.}
        \label{estimator_diagram}
    \end{figure}

     The idea of cascade architecture from DE is extended to a new estimator called ``Ballistic Multibody Estimator" (BME). Instead of using a nonlinear KF as an estimator for states of the posture, each individual $\alpha_i$ is estimated by the standard LTI discrete-time KF. The proposed estimating scheme can be visualized in Fig. \ref{estimator_diagram}.
     
    \subsection{Linear Posture Estimator}
    The posture estimator consists of an orientation estimator of the first link and a series of joint estimators. All of them are based on the standard LTI KF. The orientation state of the link $1$ comprises its orientation $\alpha_0$, angular velocity $\dot{\alpha}_0$, and angular acceleration $\ddot{\alpha}_0$. The joint state of the joint $i$ consists of its angular displacement $\alpha_i$, angular velocity $\dot{\alpha}_i$, and angular acceleration $\ddot{\alpha}_i$. Since both orientation state and joint state uses the same variable $\alpha$, the state vector ($\hat{\mathbf{x}}_{\alpha_i}$) for each KF can be defined in \eqref{eq:x_alpha_ai}. Similarly to \eqref{eq:x_k}, the state transition model of the posture estimator is assumed to have zero-mean Gaussian angular jerk $j_{\alpha_i}$. This results in a process model as seen in \eqref{eq:x_ai_k}.
    
    \begin{equation}\label{eq:x_alpha_ai}
    \mathbf{x}_{\alpha_i} = \begin{bmatrix}
        \alpha_i \\ \dot{\alpha}_i \\ \ddot{\alpha}_i
    \end{bmatrix}
    \end{equation}

    \begin{equation}\label{eq:x_ai_k}
    \mathbf{x}_{\alpha_i}[k] = \mathbf{F}_\alpha \mathbf{x}_{\alpha_i}[k-1] + \mathbf{G}_\alpha w_{\alpha_i}[k]
    \end{equation}
    
    Where:
    
    \begin{equation*}
        w_{\alpha_i} = j_{\alpha_i}
   ,~~ 
        \mathbf{F_\alpha} = \begin{bmatrix}
        1 & \Delta{t} & \frac{1}{2}\Delta{t^2}\\
        0 & 1 & \Delta{t}\\
        0 & 0 & 1\\
        \end{bmatrix}
        ,~~
        \mathbf{G_\alpha} = \begin{bmatrix}
        \frac{1}{6}\Delta{t}^3 \\
        \frac{1}{2}\Delta{t}^2 \\
        \Delta{t}
        \end{bmatrix}
    \end{equation*}
    
$\mathbf{F}_\alpha$ is the state-transition matrix. $\mathbf{G}_\alpha$ is the process-noise matrix. Since both $\mathbf{F}_\alpha$ and $\mathbf{G}_\alpha$ are constant, they are used in all posture estimators.

The orientation estimator uses the measurement from the gyroscope. The rest of the joint estimators use the corresponding encoders. Thus, the gyroscope output and encoder outputs can be modeled as \eqref{eq:gyro_enc}. These measurements resulted in observation matrices $\mathbf{H_{\alpha_0}}$ and $\mathbf{H_{\alpha}}$, which are the orientation of the first link, and joint measurement matrix of each joint, respectively.

\begin{equation}\label{eq:gyro_enc}
    y_{\alpha_0}[k] = \dot{\alpha}_0[k]
    ,~~
    \mathbf{y}_{\alpha_i}[k] = 
    \begin{bmatrix}
     \alpha_i[k] & \dot{\alpha}_i[k]
    \end{bmatrix}^\top
\end{equation}
    
\begin{equation}
    \mathbf{H}_{\alpha_0} = \begin{bmatrix}
    0 & 1 & 0 \\
    \end{bmatrix}
    ,~~
    \mathbf{H_{\alpha_i}} =  \begin{bmatrix}
    1 & 0 & 0 \\
    0 & 1 & 0 \\
    \end{bmatrix}
\end{equation}
  
  In conclusion, the BME consists of a collection of LTI Kalman filters that estimates the posture and a linear time-varying KF that estimates the CoM.

\section{Simulation}
The estimator is verified using Mon$\chi$, a two-link mobile acrobatic robot. Mon$\chi$ is simulated with Simscape and connected to the BME, which is implemented in Simulink. The reference trajectory ($\mathbf{q}_r$) is a segment of the robot performing a single back somersault in the air. The system was tested under two situations. First, the robot parameters are known. Second, the robot properties are altered based on bounded uncertainty from \cref{table:uncertainty}, which reflects uncertainty from the fabrication process. In the second scenario, 300 sets of the parameters in \cref{table:uncertainty} are generated and tested. Additionally, the initial states are known in both scenarios.

\begin{table}[htpb]
    \begin{center}
        \caption{\label{tab:exp_param}Robot Parameters for the Experiment}
        \label{table:uncertainty}
        \renewcommand{\arraystretch}{1.1}
        \begin{tabular}{ | m{1.6cm} | m{3.6cm} |} 
            \hline
            Parameter & Value \\ 
            \hline
            $m_1$ & $\SI{1.5790 (76)}{\kilo\gram}$ \\ 
            \hline
            $m_2$ & $\SI{1.4370 (400)}{\kilo\gram}$ \\ 
            \hline
            $r_1$ & $\SI{0.1443 (51)}{\meter}$\\ 
            \hline
            $r_2$ & $\SI{0.1268 (34)}{\meter}$ \\ 
            \hline
            $a_1$ & $\SI{-0.0055 (5)}{\meter}$\\ 
            \hline
            $a_2$ & $\SI{0.0001 (2)}{\meter}$ \\ 
            \hline
            $I_{1}$ & $\SI{0.0375 (13)}{\kilo\gram\square\meter}$ \\ 
            \hline
            $I_{2}$ & $\SI{0.0237 (9)}{\kilo\gram\square\meter}$ \\ 
            \hline
            $\beta_1$ & $\SI{0.20 (4)}{\newton\meter\per\radian\per\second}$ \\ 
            \hline
        \end{tabular}
    \end{center}
\end{table}

The experiment was conducted upon these algorithms, which consist of the BME, the EKF, the UKF, the DE with EKF ($\bar{\text{EKF}}$) and the DE with UKF ($\bar{\text{UKF}}$). The results are shown in \cref{table:results_ideal}, \cref{table:results_uncertain}, which show that $\bar{\text{EKF}}$ and $\bar{\text{UKF}}$ perform slightly better than BME in a scenario with known parameters. However, when the uncertainty arises, the BME significantly outperforms both. These demonstrate the potential of BME to be implemented in the field because these uncertainties constantly occur while fabricating each part and also when parts inevitably degrade. BME also has significantly less execution time, which is the result of less algorithm complexity. This demonstrates the potential of the algorithm as a lightweight and highly scalable state estimation algorithm \cref{table:computational time}.

\begin{figure}[h!]
    \centerline{\includegraphics[width=\textwidth/2]{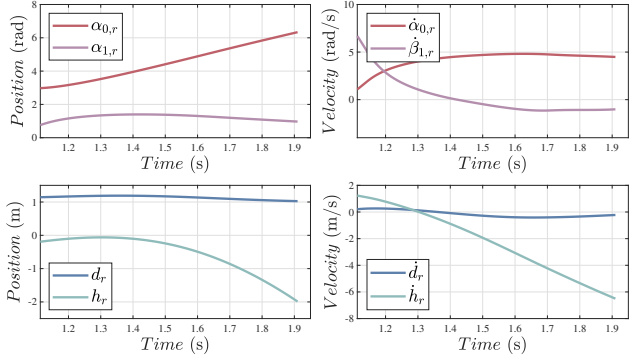}}
    \caption{Reference position and velocity trajectory.}
    \label{fig:pose_est}
\end{figure}

\begin{figure}[h!]
    \centerline{\includegraphics[width=\textwidth/2]{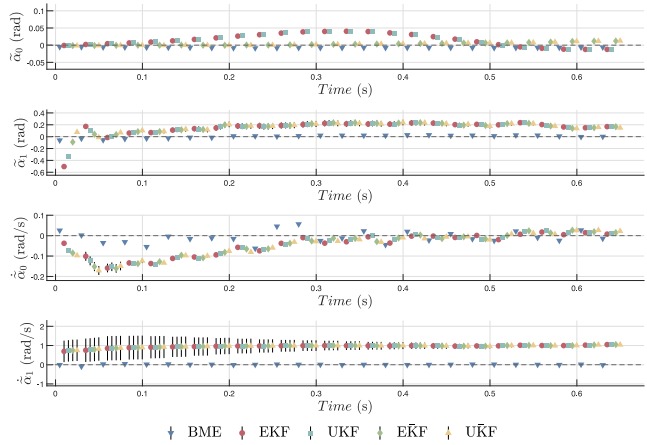}}
    \caption{A comparison between multiple algorithm in orientation and joint states estimation.}
    \label{fig:pose_est}
\end{figure}

\begin{figure}[h!]
    \centerline{\includegraphics[width=\textwidth/2]{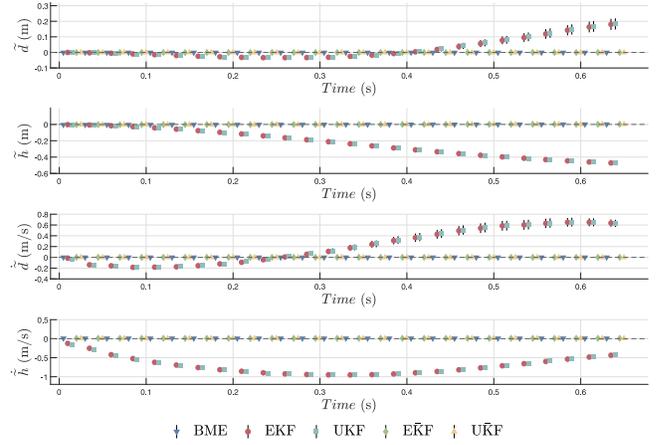}}
    \caption{A comparison between multiple algorithm in center-of-mass position and velocity estimation.}
    \label{fig:pose_est}
\end{figure}

\begin{table}[h!]
    \begin{center}
        \caption{The Average Time per Computational Step from the Simulation of Each Algorithm}
        \label{table:computational time}
        \renewcommand{\arraystretch}{1.25}
        \begin{tabular}{|c|c|c|c|c|}
        \firsthline
        BME & EKF & UKF & $\bar{\text{EKF}}$ & $\bar{\text{UKF}}$\\
        \hline
        \SI{32.05}{\micro\second} &  \SI{144.21}{\micro\second} & \SI{227.13}{\micro\second} &   \SI{61.42}{\micro\second} & \SI{75.54}{\micro\second} \\
        \lasthline
        \end{tabular}
    \end{center}
\end{table}

\begin{table}[h!]
    \begin{center}
        \caption{The Root-Mean-Square Error (RMSE) of Each Algorithm with Known Parameters}
        \label{table:results_ideal}
        \renewcommand{\arraystretch}{1.25}
        \begin{tabular}{|m{1.5cm}||c|c|c|c|c|}
        \firsthline
        RMSE & BME & EKF & UKF & $\bar{\text{EKF}}$ & $\bar{\text{UKF}}$\\
        \hline
        $\tilde{\alpha}_0~(\si{\radian})$                   & 6.10e-3 & 4.87e-3 & 4.82e-3 & 4.51e-3 & 4.51e-3\\
        $\tilde{\alpha}_1~(\si{\radian})$                   & 2.45e-2 & 9.21e-3 & 9.21e-3 & 9.22e-3  & 9.23e-3\\
        $\dot{\tilde{\alpha}}_0~(\si{\radian\per\second})$  & 8.97e-3 & 1.73e-2 & 1.70e-2 & 5.24e-3  & 5.14e-3\\
        $\dot{\tilde{\alpha}}_1~(\si{\radian\per\second})$  & 9.37e-3 & 4.78e-3 & 4.78e-3 & 4.76e-3 & 4.76e-3\\
        \hline
        $\tilde{d}~(\si{\meter})$                           & 3.80e-4 & 1.14e-2 & 1.13e-2 & 3.80e-4 & 3.80e-4\\
        $\tilde{h}~(\si{\meter})$                           & 1.96e-3 & 9.78e-2 & 9.58e-2 & 9.80e-4 & 9.90e-4\\
        $\dot{\tilde{d}}~(\si{\meter\per\second})$          & 1.22e-4 & 7.44e-3 & 7.42e-3 & 1.24e-4 & 1.21e-4\\
        $\dot{\tilde{h}}~(\si{\meter\per\second})$          & 8.24e-3 & 1.23e-2 & 1.21e-2 & 8.32e-3 & 8.31e-3\\
        \lasthline
        \end{tabular}
    \end{center}
\end{table}

\begin{table}[h!]
    \begin{center}
        \caption{The Root-Mean-Square Error (RMSE) of Each Algorithm with Uncertain Parameters}
        \label{table:results_uncertain}
        \renewcommand{\arraystretch}{1.25}
        \begin{tabular}{|m{1.5cm}||c|c|c|c|c|}
        \firsthline
        RMSE & BME & EKF & UKF & $\bar{\text{EKF}}$ & $\bar{\text{UKF}}$\\
        \hline
        $\tilde{\alpha}_0~(\si{\radian})$                       & 7.10e-3   & 2.21e-2   & 2.31e-2   & 2.31e-2   & 5.00e-3\\
        $\tilde{\alpha}_1~(\si{\radian})$                       & 2.54e-2   & 1.88e-1   & 1.83e-1   & 1.88e-1   & 1.88e-1\\
        $\dot{\tilde{\alpha}}_0~(\si{\radian\per\second})$      & 2.97e-2   & 6.98e-2   & 6.95e-2   & 6.98e-2   & 7.00e-2\\
        $\dot{\tilde{\alpha}}_1~(\si{\radian\per\second})$      & 2.32e-2   & 9.54e-1   & 9.53e-1   & 9.54e-1   & 9.53e-1\\
        \hline
        $\tilde{d}~(\si{\meter})$                               & 6.00e-4   & 6.66e-2   & 6.43e-2   & 6.00e-4   & 6.00e-4\\
        $\tilde{h}~(\si{\meter})$                               & 2.21e-3   & 2.74e-1   & 2.72e-1   & 2.10e-3   & 2.10e-3\\
        $\dot{\tilde{d}}~(\si{\meter\per\second})$              & 1.12e-4   & 3.68e-1   & 3.68e-1   & 1.10e-4   & 1.02e-4\\
        $\dot{\tilde{h}}~(\si{\meter\per\second})$              & 9.10e-3   & 7.40e-1   & 7.38e-1   & 9.12e-3   & 9.11e-3\\
        \lasthline
        \end{tabular}
    \end{center}
\end{table}

\section{Conclusion}
In this paper, we have discussed state estimation for flying open kinematic chains. Since the system is non-linear, several estimation techniques can be used, such as extended Kalman filter and unscented Kalman filter. However, from a property of the system, where its posture state is independent of the position of the center of mass, this paper proposed and implemented a cascade Kalman filter scheme for the $n$-link open kinematic chain; so-called ``Ballistic Multibody Estimator." The algorithm consists of a traditional Kalman filter that does not require the computation of Jacobian for each time step; hence, it significantly reduces the execution time of the algorithm. The simulation results also show that the ballistic multibody estimator performance is better than both extended Kalman filter and unscented Kalman filter when the robot parameters are uncertain. 

The model used for the proposed algorithm is generalized and can be extended for the closed-kinematic chain with some adjustment in sensors.

\bibliographystyle{IEEEtran}

\bibliography{root.bib}

\end{document}